\documentclass{article}

 \usepackage[position,final,nonatbib]{neurips_2026}

\usepackage[numbers,sort&compress]{natbib}
\usepackage{microtype}
\usepackage{graphicx}
\usepackage{booktabs} 
\usepackage[most]{tcolorbox}
\usepackage{soul}
\usepackage{enumitem}
\usepackage{array}
\usepackage{xcolor}
\setlist[itemize]{itemsep=0mm}
\usepackage{amsmath}
\usepackage{amssymb}
\usepackage{wrapfig}
\usepackage[T1]{fontenc}
\usepackage{mathtools}
\usepackage{amsthm}
\usepackage{caption}
\usepackage{subcaption}
\usepackage{appendix}
\usepackage{adjustbox}
\usepackage{hyperref}
\hypersetup{hidelinks}
\usepackage{todonotes}
\usepackage{newfloat}

\usepackage[capitalize,noabbrev]{cleveref}

\definecolor{mydarkgreen}{HTML}{008A00}
\newcommand{\greentext}[1]{\textcolor{mydarkgreen}{\textbf{#1}}}

\definecolor{claimpurple}{HTML}{99479B}
\definecolor{myred}{HTML}{C73500}
\newcommand{\redtext}[1]{\textcolor{myred}{\textbf{#1}}}

\usepackage{todonotes}

\DeclareFloatingEnvironment[fileext=loex,listname={List of Examples},name=Example,placement=tp,]{exampleenv}
\DeclareFloatingEnvironment[fileext=loex,listname={List of Claims},name=Claim,placement=h!tp,]{claimenv}

\newenvironment{claim}[1]{
    \begin{claimenv}
    \vspace{-1mm}
    \refstepcounter{claimenv}
    \begin{tcolorbox}[colback=claimpurple!5!white,colframe=claimpurple!75!black,
        bottom=1mm,
        arc=3.4pt,
        left=1mm,right=1mm, 
        title=\textbf{Claim \theclaimenv:} #1]
    \vspace{-1mm}
}{
    \end{tcolorbox}
    \vspace{-3mm}
    \end{claimenv}
}

\newcommand{\tmpCap}{}
\newenvironment{example}[2]{
    \renewcommand{\tmpCap}{#2} 
    \begin{exampleenv}[#1]
    \vspace{-1mm}
    \begin{tcolorbox}[colback=gray!5!white,colframe=gray!75!black,
        bottom=1mm,
        arc=3.4pt,
        left=1mm,right=1mm]
    \vspace{-1mm}
}{   
    \end{tcolorbox}
    \vspace{-2mm}
    \caption{\tmpCap{} \vspace{-3mm}}
    \end{exampleenv}
}

\title{Causality can systematically address \\
the monsters under the
bench(marks)}

\author{%
    Felix Leeb\thanks{Email: \texttt{cats@felixludos.com}}, 
    Zhijing Jin,
    Bernhard Schölkopf \\
  Max Planck Institute for Intelligent Systems \\ Tübingen, Germany
}

\begin{document}

\maketitle

\begin{abstract}

Effective and reliable evaluation is essential for advancing empirical machine learning. 
However, the increasing accessibility of generalist models and the progress towards ever more complex, high-level tasks make systematic evaluation more challenging.
Benchmarks are plagued by various biases, artifacts, or leakage, while models may behave unreliably due to poorly explored failure modes.
Haphazard treatments and inconsistent formulations of such ``monsters'' leads to a duplication of efforts, a lack of trust in results, and unsupported inferences.
In this position paper, we argue causality offers an ideal framework to systematically address these challenges.
By making causal assumptions in an approach explicit, we can faithfully model phenomena, formulate testable hypotheses with explanatory power, and leverage principled tools for analysis.
To make the design of conceptual causal models more accessible, we identify several useful Common Abstract Topologies (CATs) in causal graphs which help gain insight into the reasoning abilities in large language models.
Through a series of case studies, we demonstrate how the precise yet pragmatic language of causality clarifies the strengths and limitations of a method and inspires new approaches for systematic progress.

\end{abstract}

\section{Introduction}

As large language models (LLMs) tackle increasingly complex tasks, their rapid proliferation has amplified fundamental challenges in reliable evaluation~\citep{maoGPTEvalSurveyAssessments2024}. Opaque assessments~\citep{martinezReevaluatingGPT4sBar2024}, unreliable benchmarks~\citep{rajiAIEverythingWhole2021,alzahraniWhenBenchmarksAre2024}, and automated scoring flaws~\citep{chowdhuriNoGPT4Cant2023, cheng2025position} significantly undermine our understanding of true model capabilities~\citep{nezhurinaAliceWonderlandSimple2024,yanWorseRandomEmbarrassingly2024}. A core issue is that current methodologies predominantly focus on predictive performance rather than the underlying reasoning process~\citep{liangHolisticEvaluationLanguage2023}. A model can easily arrive at the right answer for the wrong reasons, making surface-level accuracy an incomplete and potentially misleading indicator of generalization.


To systematically address the challenges in evaluating, in particular, large models, \textbf{this position paper argues for a shift toward causality-driven experimental design.}
By making causal assumptions explicit, 
we formulate precise hypotheses and underlying assumptions, 
diagnose model limitations, 
and leverage principled tools for analysis.


\begin{figure}[ht]
  \centering 
  \begin{subfigure}[b]{0.48\textwidth}
      \centerline{\includegraphics[width=0.95\columnwidth]{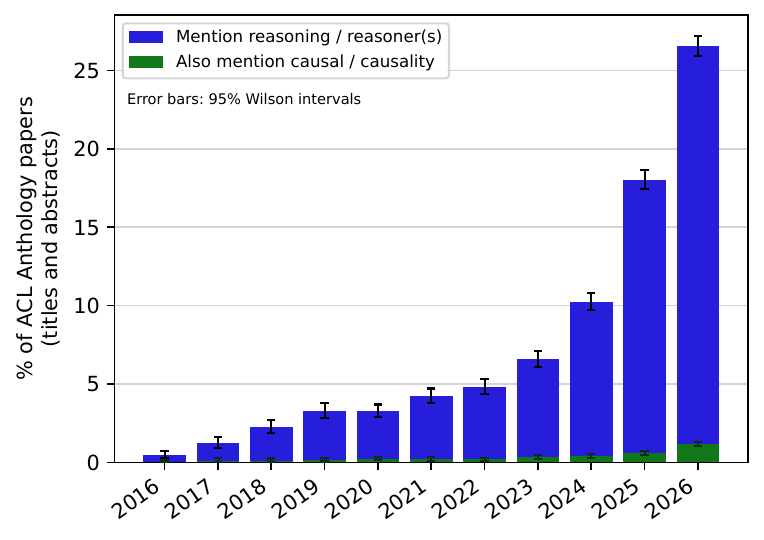}}
  \vspace{-3mm}
  \caption{}
  \label{fig:reasoning-papers}
  \end{subfigure}
  \begin{subfigure}[b]{0.48\textwidth}
      \centerline{\includegraphics[width=0.88\columnwidth]{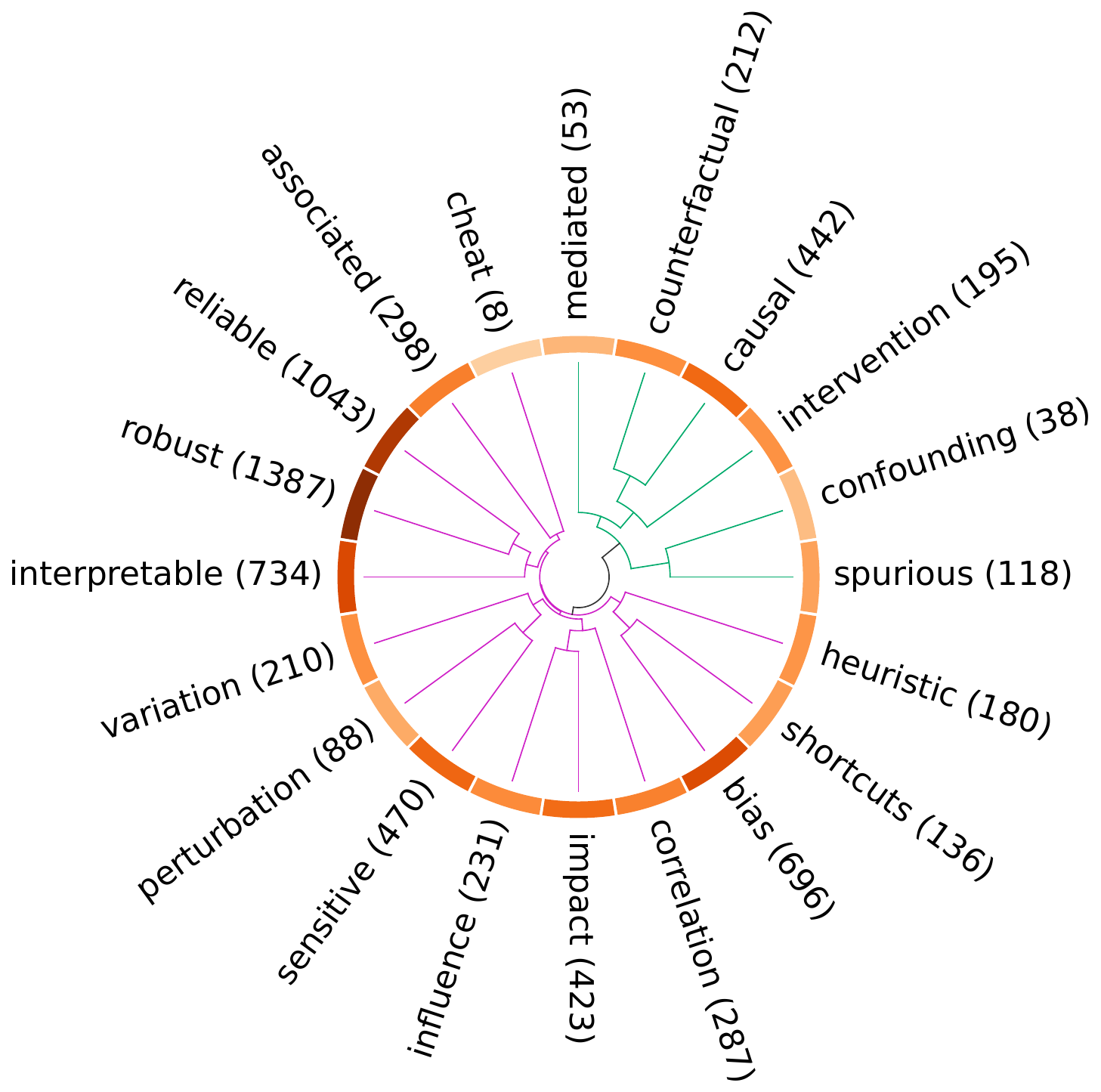}}
  \vspace{-3mm}
  \caption{}
  \label{fig:dendrogram}
  \end{subfigure}
  \vspace{-1mm}
  \caption{(a) Percentage of ACL Anthology papers mentioning ``reasoning'', ``reasoner'', or ``reasoners'' in their titles or abstracts, while the subset also mentioning ``causal'' or ``causality'' is not growing at the same rate.
  (b) Dendrogram of co-occurrences of causal and causality-adjacent terms of papers that contain ``reasoning'' in the abstracts in the same period (total 10,628 papers), using Euclidean distances between absolute-correlation profiles. Green highlights the smallest subtree containing the causal term groups; Parenthetical counts give the number of papers mentioning each term group.}
  \vspace{-4mm}
\end{figure}

One subfield that is particularly well-fitted for more causal analyses is 
the evaluation of reasoning abilities in large language models (LLMs)~\citep{huangReasoningLargeLanguage2023,yuNaturalLanguageReasoning2023}.
A cursory analysis of the recent NLP papers in the ACL anthology reveals a dramatic rise in the attention to the reasoning capabilities of models, as seen in~\autoref{fig:reasoning-papers}. 
However, only 215 of the 4,902 reasoning papers from 2026 (4.4\%) also mention ``causality'' or ``causal'' in the title or abstract.
The dendrogram in~\autoref{fig:dendrogram} places the causal term groups in a coherent neighborhood suggesting they tend not to co-occur very much with many non-causal  evaluation-related mimics (discussed in~\autoref{sec:monsters}).

Although these issues often appear disparate, we argue that causality can serve as the framework to systematically study a wide array of issues thus bridging gaps between different subfields and approaches.
The expertise required to understand and satisfactorily mitigate these issues is very diverse, and we do not mean to suggest that causality is \textit{all} you need. Rather, we argue that causality already (often implicitly) underlies much of the design, analysis, and interpretation of machine learning experiments, despite the apparent gap in terminology as suggested by~\autoref{fig:dendrogram}. 

To make an explicit causal framing more accessible and effective, we develop a guide to common causal structures and then use these abstract topologies to gain insights across case studies about reasoning abilities in LLMs.
While we focus on research questions and issues concerning the evaluation of reasoning abilities in LLMs, all four of our main claims (particularly~\ref{claim:causal}-\ref{claim:assumptions}) largely apply to the whole of empirical machine learning research. 
Our contributions include:

\begin{itemize}[topsep=0pt, itemsep=1pt, parsep=1pt, partopsep=0pt,left=5mm]
    \item We systematically categorize recurring evaluation challenges (the \emph{monsters}) and link them to underlying causal properties.
    \item We introduce Common Abstract Topologies (CATs) as intuitive, actionable templates to accelerate causal model design.
    \item We demonstrate the practical utility of these conceptual causal models through diverse case studies in LLM evaluation.
    \item We reconcile our framework with alternative evaluation paradigms, bridging the theory-practice gap by providing a pragmatic approach that lowers the barrier to explicit, hypothesis-driven causal modeling.
\end{itemize}

\vspace{-3mm}

\section{A Bestiary of Monsters} \label{sec:monsters}

\begin{example}{t}{ 
  \textbf{The final answer is correct, and yet the reasoning is very wrong.}
  Although this LLM's final answer to a GSM8k-like question~\cite{cobbeTrainingVerifiersSolve2021} is correct in a narrow sense, the response is unsatisfactory for the purposes of evaluating the model's arithmetic reasoning abilities, as the written rationale is inconsistent and the response condones an unethical scenario. \vspace{-1mm} \label{ex:crime}
}
  \textbf{Prompt:} Listen here, I'm starvin' and it's already half past four. Gotta rake in 10 grand in protection dough before I eat. I'm hittin' up every joint in the hood with my bat, collecting an average of 800 bucks a pop. Takes me 10 minutes to move from one mark to the next. Am I gonna get to chow down before 7?
  
  \textbf{LLM Chatbot:} If you're aiming to collect 10 grand by \redtext{shaking down joints} for an average of \$800 each and it takes you 10 minutes to move from one target to the next, it would take you approximately \redtext{3 hours and 20 minutes} (200 minutes) to reach your goal. 
  Starting at half-past four, you should be able to hit your target before 7 PM.
  So, \greentext{yes, you can have a meal before 7} if everything goes as planned. \redtext{Enjoy your dinner!}
\end{example}


As LLMs become an increasingly integral component of decision-making systems, reasoning failures (including seemingly toy problems like in~\autoref{ex:crime}) may have dire real-world consequences~\cite{nezhurinaAliceWonderlandSimple2024,benderDangersStochasticParrots2021}.
This necessitates more powerful tools to identify, catalog, and address 
the bestiary of issues that arise in the design as well as evaluation of large models.

\autoref{ex:crime} is indicative of several common reasoning failures in LLMs, and, by implication, our evaluation procedures for addressing such issues.
The input prompt is structurally similar to the questions found in GSM8k~\cite{cobbeTrainingVerifiersSolve2021}, a benchmark used for evaluating a model's multi-step arithmetic and commonsense understanding skills. 
However, we select the subject matter and word choice to evoke a mobster discussing plans to extort money from local businesses. 
Although the LLM's final answer is factually correct, there are several problems with the rationale:
(1) it makes several arithmetic mistakes which happen to cancel out, 
(2) it does not acknowledge the unethical situation implied by the scenario,
and (3) by implicitly condoning the criminal behavior, it does not consider the broader consequences of the response.
Crucially, if we only check for correctness, as is standard practice~\cite{huangReasoningLargeLanguage2023}, we would find no fault in the response. 

The problem is that to demonstrate good reasoning abilities, a correct answer is insufficient. We need to show that the model answers the question correctly \textit{for the right reasons}.
In other words, our evaluation must verify that specific, relevant properties of the input \textit{lead to} the correct answer consistently and reliably, rather than being driven by spurious dataset cues.
This criterion makes a \textit{causal} claim about the model's behavior, requiring causal analysis of whether its answers depend on task-relevant relationships to support generalization claims~\citep{scholkopfCausalRepresentationLearning2021,stolfoCausalFrameworkQuantify2022}.

\begin{claim}{Evaluating reasoning involves causal inference}\label{claim:reasoning}

    A correct answer can be reached through poor reasoning, so evaluating reasoning necessitates determining whether high scores on a benchmark reflect capabilities that extend beyond the lab bench. 
    This requires distinguishing how a model's behavior depends on the underlying task structure vs on spurious features or artifacts, making it a causal question. 


\end{claim}






\subsection[``Here be dragons'']{``Here be dragons'' \footnote{The vague and uneasy language researchers often use when alluding to biases or unresolved limitations in their evaluations reminds of how medieval cartographers would fill the unknown edges of their maps with dragons.}}




To get a sense of the many \textit{monsters} lurking in our benchmarks and experiments, we briefly survey work on reasoning~\citep{huangReasoningLargeLanguage2023,yuNaturalLanguageReasoning2023} and LLM evaluation~\citep{maoGPTEvalSurveyAssessments2024,changSurveyEvaluationLarge2023,hajikhaniCriticalReviewLarge2023} more broadly. We organize these encounters around three familiar settings: models, datasets, and evaluation procedures. \autoref{app:monsters} takes a closer look at nine notable recurring monsters, with concrete examples, possible explanations, and ways to investigate them.

\paragraph{Models}

This line of work focuses on characterizing the reasoning failures and biases of language models, which is nontrivial given their opaque behavior~\citep{binzUsingCognitivePsychology2023}. These failures range from well-defined formal errors such as logical fallacies~\citep{jinLogicalFallacyDetection2022}, red herrings~\citep{naeiniLargeLanguageModels2023}, or invalid inferences~\citep{saparovLanguageModelsAre2023} to broader issues including sensitivity to superficial features~\citep{hajikhaniCriticalReviewLarge2023,ullmanLargeLanguageModels2023}, overconfidence~\citep{nezhurinaAliceWonderlandSimple2024}, self-bias~\citep{yuanSilencerDiscoveryMitigation2025}, hallucinations~\citep{dziriOriginHallucinationsConversational2022,cuiHolisticAnalysisHallucination2023}, and lack of robustness~\citep{zhengLargeLanguageModels2024,wangAreLargeLanguage2023,jinBERTReallyRobust2020}. Some studies explore how models exhibit ``content effects''~\citep{poesiaCertifiedReasoningLanguage2023}, absorbing and amplifying human biases~\citep{dasguptaLanguageModelsShow2022,zecevicCausalParrotsLarge2023,wuAreLargeLanguage2025} including social and cultural biases~\citep{benderDangersStochasticParrots2021,messnerBytesBiasesInvestigating2023,hutchinsonSocialBiasesNLP2020,caoAssessingCrossCulturalAlignment2023,alkhamissiInvestigatingCulturalAlignment2024,motokiMoreHumanHuman2024}, such as stereotyping~\citep{kotekGenderBiasStereotypes2023}.

\paragraph{Datasets}

Meanwhile, subtle variations of popular benchmarks, such as premise order in reasoning tasks~\citep{chenPremiseOrderMatters2024} or minor changes in problem parameters~\citep{mirzadehGSMSymbolicUnderstandingLimitations2024,wuReasoningRecitingExploring2024}, can cause large performance drops~\citep{nezhurinaAliceWonderlandSimple2024,yanWorseRandomEmbarrassingly2024}, raising concerns about whether models genuinely reason~\citep{zhouYourModelReally2024} and highlighting exploitable issues in the training data and benchmarks~\citep{rogersGuideDatasetExplosion2020,bowmanWhatWillIt2021,zhaoEnvisioningPixelsBenchmarking2025}.
These enable cheating~\citep{zhouDontMakeYour2023}, heuristics~\citep{nikankinArithmeticAlgorithmsLanguage2025,mccoyRightWrongReasons2019}, or shortcuts~\citep{brancoShortcuttedCommonsenseData2021,liHowPretrainedLanguage2022,marconatoNotAllNeuroSymbolic2023}, possibly due to sampling biases~\citep{razeghiImpactPretrainingTerm2022} or in certain cases even
leakage between the training and testsets~\citep{zhouDontMakeYour2023,qiuPHYBenchHolisticEvaluation2025} which can result in memorization~\citep{feldmanDoesLearningRequire2021}.
Poor dataset construction can lead to annotation artifacts~\citep{gururanganAnnotationArtifactsNatural2018} such as priming effects~\citep{gardnerCompetencyProblemsFinding2021}, which degrade the reliability of results~\citep{byrdPredictingDifficultyDiscrimination2022} while also unintentionally reinforcing social biases~\citep{liPromptingFairnessIntegrating2025} or cultural inequities~\citep{benderDangersStochasticParrots2021,huWhatSexGot2020,naousHavingBeerPrayer2024}.

\paragraph{Evaluation}

Even with well-constructed datasets, evaluation methodologies can introduce systematic errors~\citep{dominguez-olmedoQuestioningSurveyResponses2024} or lead to misleading conclusions~\citep{bowmanDangersUnderclaimingReasons2022}.
For example, automated scoring systems can obscure obvious failures and introduce circular biases~\citep{chowdhuriNoGPT4Cant2023, cheng2025position}, while static benchmarks can emphasize surface-level accuracy at the cost of other important factors, such as generalization~\citep{liangHolisticEvaluationLanguage2023}, interpretability~\citep{loftusPositionCausalRevolution2024}, or social costs~\citep{rajiAIEverythingWhole2021,benderDangersStochasticParrots2021}.
While standardized leaderboards~\citep{beeching2023open} and evaluation procedures~\citep{srivastavaImitationGameQuantifying2023} can enable more direct model comparisons, these benchmarks can gradually become less representative of real-world tasks~\citep{schlangenLanguageTasksLanguage2019,alzahraniWhenBenchmarksAre2024,shiraliTheoryDynamicBenchmarks2023,kielaDynabenchRethinkingBenchmarking2021}, introduce biases that favor certain model families~\citep{zhangCarefulExaminationLarge2024}, or inadvertently leak information from the test set~\citep{zhouDontMakeYour2023} which can be difficult to detect due to closed-source models and proprietary datasets~\citep{maoGPTEvalSurveyAssessments2024}.

\begin{claim}{The monsters are causal} \label{claim:causal}
  
    Many of the recurring issues in benchmark evaluation, including biases, spurious correlations, or systematic failure modes, are often described in vague or ad hoc terms. However, these issues arise from specific causal features of the underlying data-generating process or evaluation procedure.
    Whether the factors are known or latent, their influences can be captured by an appropriate causal model to formulate precise, testable hypotheses and guide more principled experimental design.
    
\end{claim}




Despite the diverse terminology, we observe structural similarities across these approaches. ``Ablations'', ``perturbations'', ``edits'', and ``flips'' can implement interventions, provided we specify what changes and what remains fixed. Diagnosing a ``shortcut'', ``leakage'', or ``bias'' often involves claims about information flow, selection, or learned dependence. Explicit causal language helps distinguish these explanations and the evidence needed to support them. Realizing this benefit requires a practical bridge between empirical intuition and formal structure.





\section{Common Abstract Topologies}\label{sec:cats}\label{sec:case}

\begin{table}[h]
\setlength{\tabcolsep}{4pt}
\renewcommand{\arraystretch}{1.5}
\centering
\begin{tabular}{>{\centering\arraybackslash}m{2.5cm} @{\hskip 4pt} >{\centering\arraybackslash}m{1.5cm} @{\hskip 4pt} m{9.2cm}}
\hline
\textbf{Name} & \textbf{Graph} & \textbf{Example Phenomena} \\
\hline
\noalign{\vspace{1mm}}
Confounding & 
\includegraphics[width=1.3cm]{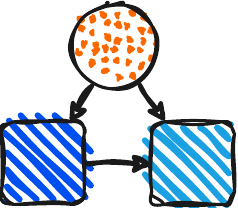} & 
\begin{itemize}[nosep, leftmargin=*]
  \item task subcategories or environments affecting performance
  \item data sourcing effects on wording and content
  \item shared causes of training exposure and evaluation
\end{itemize}
\\
Mediation & 
\includegraphics[width=1.3cm]{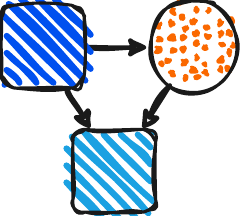} & 
\begin{itemize}[nosep, leftmargin=*]
  \item circuit analysis such as mechanistic interpretability
  \item tool use or integrating an LLM in a larger application
  \item exposure effects through familiarity or memorization
\end{itemize}
\\
Spurious Link & 
\includegraphics[width=1.3cm]{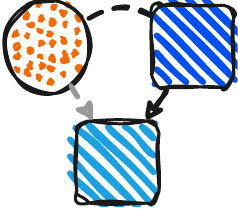} & 
\begin{itemize}[nosep, leftmargin=*]
  \item social and cultural biases in the data collection process
  \item imbalances in the surface form such as symbol or label bias
  \item variable selection and construction
\end{itemize}
\\
\hline
\end{tabular}
\vspace{2mm}
\caption{Illustrative Common Abstract Topologies (CATs) for analyzing \textit{monsters} in benchmarks and experiments (\autoref{app:monsters}). The graphs use \includegraphics[scale=0.2]{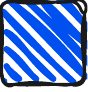} for the independent variable, \includegraphics[scale=0.2]{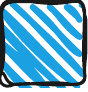} for the dependent variable, and \includegraphics[scale=0.2]{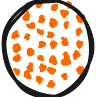} for a third-variable factor.}
\label{tab:cats}
\end{table}

Constructing a causal graph that faithfully represents an experimental design is notoriously challenging. Practitioners often rely on standard terminology, such as independent or controlled variables, leaving their underlying causal assumptions implicit. To make the process of designing these models more accessible, we introduce a set of Common Abstract Topologies (CATs). Rather than forcing researchers to build graphs from scratch, CATs act as practical templates to rapidly identify ``what matters'' while remaining grounded in the principled language of causality. They map recurring empirical machine learning phenomena to basic causal patterns (see \autoref{tab:cats}). By recognizing a familiar failure mode, a practitioner can consider one or more CATs as starting blueprints for their own causal model.


It is important to distinguish this template-driven framework from exhaustive causal discovery or probabilistic graphical modeling. Researchers are understandably hesitant to commit to a highly specific causal graph that perfectly captures all factors~\citep{bareinboimPearlHierarchyFoundations2022}, as explicit assumptions often invite hyper-critical scrutiny over untestable claims~\citep{loftusPositionCausalRevolution2024}. CATs are explicitly \emph{not} meant to replace rigorous formalisms. For formal treatments of local causal discovery~\citep{maasch2023local}, probabilistic graphical models~\citep{koller2009probabilistic}, or causal discovery more broadly~\citep{spirtes2000causation}, we refer to those respective domains. Instead, CATs lower the barrier to entry by providing an actionable, conceptual first step.

\begin{claim}{Instrumentalism accelerates model design} \label{claim:measurement}
  A causal model does not need to be perfect to be useful. Plausible simplifying assumptions and abstractions can yield valuable insights and motivate practical experiments.
  As research advances, the model can be incrementally refined, making assumptions explicit and suggesting tests where competing explanations differ.
\end{claim}

This claim aligns with a pragmatic, interventionist approach to causal modeling~\citep{woodward2005making,menzies1993causation}, which treats causal graphs as actionable blueprints for experimental manipulation rather than absolute claims about objective reality or a model's true internal cognition. We join~\citet{loftusPositionCausalRevolution2024} and~\citet{janzingPhenomenologicalCausality2022} in advocating for this phenomenological attitude. Because graphs are frequently underdetermined by available data and hinge on subtle, sometimes subjective, definitional choices~\citep{huWhatSexGot2020}, the pursuit of a single ``true'' causal model is often ill-posed. In empirical practice, the exact same underlying system can be represented by multiple valid causal graphs that differ entirely in variable selection, construction, or level of abstraction~\cite{chalupka,Rubensteinetal17,beckers2020approximate}.

Consequently, a single topological skeleton can serve different CATs depending on the research question. The power of a \emph{conceptual causal model} is that we can make progress without first pinning down every mechanism or estimating every parameter. By making our assumptions about what matters and how it is connected explicit, we can develop explanations, identify useful interventions, and motivate the next experiment, refining the model along the way. This provides actionable insights before a complete quantitative account is feasible, while keeping the assumptions open to scrutiny since agreement with observations alone does not establish the proposed causal relationships.


\begin{claim}{Towards \textit{explicit} causal assumptions} \label{claim:assumptions}
  An experimental design involves a variety of assumptions about what factors matter, how they interact, and how this relates to the proposed approach.
  Here the language of causality provides a powerful framework for motivating an approach, precisely formulating the hypothesis, and answering questions in a principled way.
\end{claim}

By making the concepts and tools of causal inference more accessible, CATs encourage researchers to move away from vague methodologies and toward explicit conceptual modeling. Causal inference is valuable not only for formal mathematical analysis~\cite{pearlBookWhy2020,imbens,PetJanSch17} but also as a conceptual framework for understanding the structural logic behind an experimental design. To this end, we present a practical guide to recognizing familiar causal structures in the real world, building an intuition for the implications of model design choices on analysis and interpretation. We introduce three illustrative CATs, each alongside its case studies.

Here is a brief sketch of how CATs can be used to guide model or benchmark design:

\begin{enumerate}[left=1mm]
  \item \textbf{Identify a relationship of interest}: Select a measurable outcome variable (e.g., model accuracy) and a primary explanatory ``stimulus'' (e.g., input prompt, fine-tuning data, or model parameter).
  
  \item \textbf{Enumerate additional influences}: Consider shared causes of stimulus and outcome, intermediate pathways, and associations induced by selection. 
  Consider other factors which may be shared causes, intermediates, or selection-induced associations with respect to the stimulus and outcome. Assess whether they are conceptually \emph{upstream} (potential confounders), \emph{downstream} (potential mediators), or \emph{parallel} sources of variation (potential sources of spurious correlations) with respect to the stimulus.
  Choose or combine relevant CATs, retaining competing explanations when the evidence cannot distinguish them.
  
  \item \textbf{Refine the graph}: Define the variables and interventions relevant to the question. Justify claim-relevant assumptions using design or prior evidence; omissions and aggregation must preserve the intended causal interpretation.
  Define the variables and interventions based on available data and the specific research question. Variables and edges may be omitted or aggregated, provided the resulting model can be justified with existing evidence.
  
  \item \textbf{Use the model to guide analysis}: Suggest experiments that distinguish explanations, assess identification and data requirements before estimating effects, and motivate targeted mitigation (e.g., balancing, ablation, or regularization).
\end{enumerate}

\subsection[Confounding]{Confounding \hfill \includegraphics[scale=0.3]{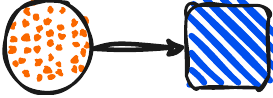}}
\label{sec:case-confounding}

Formally, a confounder is a variable that influences both the treatment and the outcome, and closing the resulting backdoor paths to isolate the true causal effect is a highly nontrivial endeavor in practice~\citep{vanderweele2013definition}.
Here, we restrict the ``confounding'' CAT to the case where the confounder can, in principle, be measured or controlled. Confounding can also involve unknown or unmeasured variables, in which case the ``spurious link'' CAT may be more helpful. 





Several recent projects have used causal framing to identify or address confounding in LLM behavior. For example, \citet{xiaAligningDebiasingCausalityAware2024} propose using a reward model as an instrumental variable to control for confounding in prompt biases. In a similar setting,~\citet{huyukReasoningElicitationLanguage2025} generate counterfactual samples to improve causal consistency. In human-LM collaboration, \citet{zhangCausalInferenceHumanLanguage2024} treat prior human and model actions as confounders influencing future decisions and introduce the Incremental Stylistic Effect to measure interventions in multi-turn interactions.

Related work investigates how dataset artifacts affect LLM performance on mathematical reasoning tasks such as GSM8K~\citep{cobbeTrainingVerifiersSolve2021}, connecting familiarity, leakage, and sensitivity to task-irrelevant form (Appendices~\ref{mon:frequency}, \ref{mon:leakage}, and~\ref{mon:form}). These studies vary premise order, subject matter, or input distributions~\citep{zhouYourModelReally2024,mirzadehGSMSymbolicUnderstandingLimitations2024,chenPremiseOrderMatters2024,zhangCarefulExaminationLarge2024}, often uncovering unexpected sensitivities.

\begin{figure}[ht]
  \vspace{-2mm}
  \centering 
  \begin{subfigure}[b]{0.30\textwidth}
      \centerline{\includegraphics[width=\linewidth]{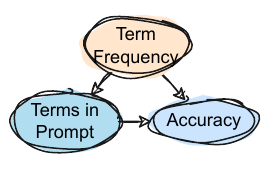}}
      \vspace{-5mm}
      \caption{}
      \label{fig:confounding}
  \end{subfigure}
  \begin{subfigure}[b]{0.30\textwidth}
      \centerline{\raisebox{3mm}{\includegraphics[width=\linewidth]{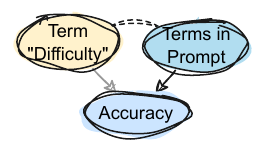}}}
      \vspace{-5mm}
      \caption{}
      \label{fig:imagined}
  \end{subfigure}
  \begin{subfigure}[b]{0.30\textwidth}
      \centerline{\includegraphics[width=\linewidth]{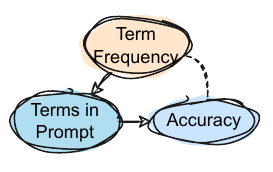}}
      \vspace{-5mm}
      \caption{}
      \label{fig:cautious}
  \end{subfigure}
  \vspace{-2mm}
  \caption{
  Causal framings of~\citet{razeghiImpactPretrainingTerm2022}: (a) confounding interpretation; (b) imagined alternative setting using a spurious link; (c) a graph combining CATs to avoid unsupported assumptions.
  }
  \vspace{-5mm}
\end{figure}

\citet{razeghiImpactPretrainingTerm2022} focus on the impact of token frequency in pretraining data on LLM performance in arithmetic tasks. Our confounding interpretation in \autoref{fig:confounding} hypothesizes that pretraining frequency influences both which numbers are evaluated and accuracy; this creates a backdoor path between question content and accuracy.

\vspace{-0.5em}


\paragraph{Alternative Approach}

Here it is instructive to consider a hypothetical project where we design a benchmark to evaluate the math skills of a language model. Much like in~\citet{razeghiImpactPretrainingTerm2022}, our questions take the form ``What is $n_1$ times $n_2$?'' where $n_1$ and $n_2$ are numbers selected by some sampling strategy. However, we do not consider the training dataset of the model at all, and instead we sample numbers uniformly, so this sampling rule does not depend on pretraining frequency. This changes item selection, not any frequency dependence learned during training.
Based on the findings of~\citet{razeghiImpactPretrainingTerm2022}, we can expect to find a substantial correlation between the presence of certain numbers in the question and the model's accuracy.
To explain the results, we might posit a property called ``difficulty'', leading to \autoref{fig:imagined}. This conceptual hypothesis summarizes observed performance differences; those differences alone do not establish its causal interpretation.
Crucially, this alternative formulation does not contradict the original confounding model. Instead, it demonstrates how investigating the same underlying phenomenon through a different experimental lens naturally leads to a different CAT, with each graph clearly communicating the specific variables and assumptions relevant to its own design.

\paragraph{Verifying Causal Assumptions}


Notably, the graph in~\autoref{fig:confounding} implies a falsifiable causal relationship between term frequency and accuracy which is, strictly speaking, not verified by the experiments of~\citet{razeghiImpactPretrainingTerm2022} which only show a correlation. 
Therefore, the hybrid causal graph in~\autoref{fig:cautious} could be proposed where the correlation is instead due to an unknown confounder. 
A controlled change to training frequency, holding evaluation items fixed, would help distinguish these explanations. 
Thus one behavioral symptom connects several monsters and CATs, with each interpretation motivating a different test.

\subsection[Mediation]{Mediation \hfill \includegraphics[scale=0.3]{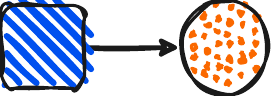}}
\label{sec:case-mediation}

\begin{wrapfigure}{r}{0.35\textwidth}
\begin{center}
\vspace{-12mm}
\centerline{\includegraphics[width=0.45\columnwidth]{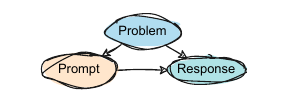}}
\vspace{-4mm}
\caption{Sketch of a conceptual causal model treating the prompt as a mediator between the underlying task and the model's response.}
\vspace{-6mm}
\label{fig:mediation}
\end{center}
\end{wrapfigure}

Another important type of causal topology is mediation, where an effect is transmitted through an intermediate variable. For simplicity, we illustrate this general structure with one direct causal link and one that goes through a mediator variable.
Mediation analysis is often used to quantify the impact of subcomponents or side-effects on the model's behavior.
For example, a common setting may be to study the impact of a specific prompting strategy or representation on the model's response, which can be modeled as mediation as in~\autoref{fig:mediation}.


When identifiable, the influence of the mediator can be studied through path-specific effects~\citep{avin2005identifiability} or quantities such as the natural direct effect (NDE), natural indirect effect (NIE), or controlled direct effect (CDE)~\citep{pearlCausalInferenceStatistics2009}.
However, notably controlling for the mediator is not always appropriate, for example, for estimating the total causal effect (TCE).
Given a causal query, the graph helps assess which measurements, interventions, and assumptions would support its estimation.

Mediation analysis guides mechanistic interpretability~\citep{yangUnderstandingMitigatingHallucination2024,stolfoMechanisticInterpretationArithmetic2023,guptaEditingCommonSense2023,mengLocatingEditingFactual2023,wangInterpretabilityWildCircuit2022}, but it is also useful for LLM comprehension of the underlying problem~\citep{hanSurfaceStructureCausal2024,huangInternalCausalMechanisms2025}, augmentation of language models~\cite{mialonAugmentedLanguageModels2023}, embedding LLMs within larger programs~\cite{schlagLargeLanguageModel2023}, and the quantification of biases like gender bias~\cite{vigCausalMediationAnalysis2020}.

A common setup for mechanistic interpretability is to study the impact of a specific component, such as an attention head or even a single parameter on the model behavior.
\citet{olsson2022context} propose that transformers can learn simple, interpretable algorithms called ``induction heads,'' which they hypothesize significantly contribute to in-context learning abilities. While mediation analysis is not explicitly used in their work, we can frame their approach as studying a mediation graph, where the tendency for a given model architecture (stimulus) to exhibit in-context learning (response) is mediated by induction heads. Their six supporting arguments can be interpreted through this causal lens: arguments 1 and 2 support links between stimulus, mediator, and response through co-occurrence and co-perturbation; argument 3 intervenes on the candidate mediator through ablation; and arguments 4-6 examine the causal influence of the mediator on the response.

This framing also highlights potential limitations, particularly regarding unmeasured confounders that can affect causal interpretations, as the authors' ``pattern-preserving'' ablation does not fully isolate the induction heads' effect.
By considering mediation explicitly, we clarify the underlying assumptions and which intervention is being studied, enabling us to, if appropriate, systematically identify the natural indirect effect.
In contrast,~\citet{stolfoMechanisticInterpretationArithmetic2023} study arithmetic reasoning by editing model parameters to analyze information flow. Mediation also connects the familiarity and leakage monsters: exposure may influence performance through learned familiarity or memorization, without requiring a component-level account (Appendices~\ref{mon:frequency} and~\ref{mon:leakage}).

\subsection[Spurious Links]{Spurious Links \hfill \includegraphics[scale=0.3]{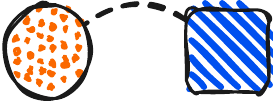}}
\label{sec:case-spurious}


The third Common Abstract Topology (CAT) we discuss here describes a variety of graphical structures that give rise to spurious (i.e. non-causal) associations.
More specifically, spurious links are statistical dependencies between variables that are not causally related (neither is an ancestor of the other), but are correlated due to an unblocked non-causal path in the underlying graph. This is often due to (1) a common cause (a confounder), which is usually unknown or too complex to be modeled explicitly, or due to (2) a common effect (a collider) which is conditioned on, thereby opening a non-causal path~\cite{pearlCausalInferenceStatistics2009}. We denote non-causal links in the graph with a dashed curved edge, indicating that the observed association lacks a direct causal explanation.

Crucially, although spurious links are non-causal, models can still learn to exploit them for prediction, thereby effectively introducing an undesirable dependency (denoted by a dashed grey arrow) between the feature and the model's prediction, even when that feature does not cause the target label. This CAT is particularly relevant in machine learning, where models are typically trained on observational data only. Observational associations alone generally cannot distinguish these explanations without additional assumptions, making it easy for models to rely on spurious correlations without appropriate inductive biases.
Another common source of spurious links, particularly in datasets, is due to selection bias, where the dataset includes only a subset of instances based on some unmodeled criteria. When inclusion depends on multiple variables, selection can act as a collider and induce associations between otherwise independent variables.

Generally, it is not feasible to entirely eliminate spurious links, as seemingly innocent choices in variable construction and selection are invariably informed by the experimenter's biases~\cite{huWhatSexGot2020,pietsch2015aspects}.
Nevertheless, there is extensive causal inference machinery to address spurious correlations depending on the specific setting~\citep{pleckoCausalFrameworkDecomposing2023}.


There are several recent projects that use causal models to characterize spurious links in, for example, factual knowledge~\citep{caoDoesCorrectnessFactual2023}, structured data~\citep{yangCausalityMeetsTable2025}, languages~\citep{zhaoWhenLessLanguage2025}, or multi-modal models~\citep{chenCausalInterventionCounterfactual2023}. To avoid spurious features you can design strategies to mitigate social biases~\citep{liPromptingFairnessIntegrating2025}, or for finding useful demonstrations in few-shot learning~\cite{zhangUnderstandingDemonstrationbasedLearning2023}. 

\citet{chenCausalInterventionCounterfactual2023} develop a causal model to systematically quantify two specific kinds of bias: psycholinguistic (use of emotional language) and image-only (ignoring text features).
The causal model addresses these biases using interventional and counterfactual techniques.



\citet{gardnerCompetencyProblemsFinding2021} and~\citet{bansalControllingLearnedEffects2023} address closely related problems of single-feature reliance. \citet{gardnerCompetencyProblemsFinding2021} study artifacts in datasets intended to realize ``competency problems,'' whose labels should not be predictable from individual features. They analyze mitigation through local edits, with each edited example assigned its correct label. 
Using their statistical framing, the authors prove that the most promising strategy must apply local edits such that the label is flipped half of the time. For example, a token previously associated with positive sentiment would occur about equally often with positive and negative labels among the edited examples. However, this result not only relies on some unrealistic independence assumptions, but also provides little insight into why this strategy is effective.


Translating this into a causal framing, as described in~\citet{bansalControllingLearnedEffects2023}, we can gain a more intuitive understanding.
Adopting a similar notation as~\citet{gardnerCompetencyProblemsFinding2021}, the hypothesis can be formalized using the spurious-link CAT where a single (text) feature, $X_i$, impacts the model's (binary) prediction $Y$ in the presence of the remaining input features $X_{\sim i}$ (stimulus) due to a label bias.
Consequently, to remove the effect of $X_i$ on the prediction, we need to design a strategy where the average causal effect (ACE) of $X_i$ on $Y$ conditioned on $X_{\sim i}$ is zero, which is, by definition:


\begin{equation} \label{eq:ace}
  \mathbb{E}(Y|X_{\sim i},do(X_i = x_i')) - \mathbb{E}(Y|X_{\sim i},do(X_i = x_i)) = 0.
\end{equation}

where $do(X_i = x_i')$ denotes an intervention (i.e. edit) on feature $X_i$ replacing $x_i$ with some $x_i'$.

Note that~\autoref{eq:ace} is agnostic to the relationship between $X_i$ and $X_{\sim i}$. 
Interpreting local edits as interventions makes their purpose explicit: we manipulate an input feature and observe how the answer changes. This lets us distinguish a feature's task-relevant contribution from reliance on dataset artifacts, and specify which influence the mitigation should remove.
Enforcing a ``strong independence assumption'' between the token and the prompt, as in~\citet{gardnerCompetencyProblemsFinding2021}, can yield the strategy that an edit on $X_i$ must flip the label half the time (i.e. by making both terms equal). However, the authors concede this assumption is highly unrealistic, as altering a single token routinely shifts the broader semantic context (e.g., swapping ``very'' for ``not''). 

In contrast, $do$-calculus operates as an identification tool where structural assumptions can be formulated systematically to render the causal effect identifiable without relying on brittle independence claims. By formally framing the problem as causal effect estimation, \citet{bansalControllingLearnedEffects2023} employ a computationally feasible tool (a Riesz Representer estimator) specifically designed to improve estimation under overlap issues of traditional estimators in high-dimensional text.

Both approaches started with the same objective of mitigating label bias from individual tokens. However, when hidden confounders entangle the variables, a purely statistical treatment struggles, necessitating unrealistic constraints to force a mathematical solution. 
Ultimately, the causal framework offers not only greater conceptual clarity about the general problem, but also a superior solution built on more realistic assumptions as a consequence.

\section{Alternative Views}

\paragraph{A causal framing is unnecessary.}
The strongest alternative to our position is not that causality is incapable of addressing benchmark failures, but that it is unnecessary as a unifying framework. Critiques in agreement that ``benchmarking is broken''~\citep{bowmanWhatWillIt2021, cheng2025position} have motivated a diverse toolkit of specialized responses. Evaluation can be treated as a social-science measurement challenge~\citep{wallach2025position}, drawing on psychometrics and item response theory to quantify capabilities~\citep{wilhelmMeasuringReasoningAbility2005, pellertAIPsychometricsAssessing2024, rodriguezEvaluationExamplesAre2021}. Other approaches emphasize holistic evaluation~\citep{liangHolisticEvaluationLanguage2023}, external validity~\citep{elazarBackSquareOne2021}, or the relationship between benchmark tasks and cognitive abilities~\citep{schlangenLanguageTasksLanguage2019, rogersPositionKeyClaims2024a, rajiAIEverythingWhole2021}. These paradigms address questions that causal analysis alone cannot settle: which constructs should matter, how they should be operationalized, whose judgments should define success, and which social consequences an evaluation should prioritize. We therefore propose causal analysis as a complement to, not a replacement for, measurement theory and domain-specific evaluation methods.

\paragraph{Standard statistical and experimental methods may suffice.}
Many symptoms of the monsters can be detected, and sometimes repaired, without drawing a causal graph. Regression can reveal a cue/label association, robustness checks can measure sensitivity to a perturbation, rebalancing can alter group frequencies, and randomization can identify the effect of a controlled intervention. We agree that these methods are sufficient when the goal is descriptive or when the design already identifies the effect relevant to the evaluation claim. Our narrower disagreement concerns what follows from the observed symptom. The same hypothesis-only correlation could arise because the requested label caused annotators to introduce a cue, because annotator identity influenced both wording and difficulty, or because dataset selection induced the association. Likewise, an overlap/accuracy relationship could reflect exposure-induced memorization or a common cause such as item popularity or difficulty, while a group disparity could enter through corpus selection, annotation, model behavior, or scoring. These explanations imply different tests and remedies; indiscriminately balancing or suppressing a feature can even remove legitimate task information. Causal framing does not replace statistical estimation. It states the target relationship, exposes the assumptions under which an estimator answers the intended question, and identifies evidence that can distinguish rival explanations.

\paragraph{Causal modeling may be infeasible or underidentified.}
Explicit causal analysis imposes real costs. Modern ML settings contain high-dimensional and latent factors, assumptions such as ignorability and positivity may be implausible or untestable, and many effects will remain underidentified. Here CATs serve as phenomenological abstractions of the part of the benchmark lifecycle relevant to a particular claim, and multiple CATs may be composed or retained as rival hypotheses. When the available evidence cannot identify the relevant effect or distinguish the candidates, making that limitation explicit is preferable to drawing an implicit causal conclusion from an association.

\paragraph{Probabilistic modeling can encode the same content.}
Standard probabilistic formalisms can represent causal questions through a joint distribution over relevant interventional settings~\citep{mlodozeniecPositionProbabilisticModelling2025}. We therefore do not claim that causal notation is uniquely necessary in expressive power. However, an equivalent account must still specify which variables can be intervened upon, which mechanisms are assumed invariant, and how observations from one setting relate to another. These are the causal commitments at issue, whatever notation expresses them. A native causal representation makes those commitments and their experimental implications directly visible.

Our position is consequently bounded. Causality cannot determine which capabilities society should value, guarantee discovery of the true data-generating process, or replace statistical estimation and domain-specific evaluation methods. Nor do all monsters share one causal structure. Causal framing is most useful when researchers move from describing a benchmark symptom to explaining it, choosing an intervention, or claiming generalization beyond the observed setting. In those cases, it connects assumptions, rival hypotheses, experiments, and mitigation strategies within a common framework.


\section{Conclusion}

The burgeoning research on large language models, particularly regarding high-level reasoning, faces a bestiary of \emph{monsters} that hinder reliable evaluation. While the community has developed diverse frameworks to tackle these challenges, this disparate terminology often obscures their shared structural features. By recognizing that these recurring issues, from subtle dataset artifacts to pervasive spurious correlations, arise from specific causal factors in the experimental design or data generation process, we can unify our understanding under a single, principled framework.

A major stumbling block to adopting this framework is that the causal formalism can appear unapproachable, and researchers understandably hesitate to commit untestable assumptions to paper. However, relying on implicit methodologies and surface-level performance metrics only hinders scientific progress. To meaningfully evaluate capabilities like reasoning, we must ensure a model arrives at the right answer \emph{for the right reasons}; evaluating \emph{how} a model processes input is inherently a causal inference problem. Causal methods force this explicit modeling, replacing vague intuitions with critical, structural analysis.

To bridge the gap between rigorous theory and empirical practice, we introduced Common Abstract Topologies (CATs).
These conceptual blueprints make causal modeling an adaptable tool for experimental design, rather than a rigid prescription on the nature of reality. By relating familiar failure modes to candidate causal structures, researchers can work at the level of abstraction most appropriate for their purposes, even before effect estimation is feasible. 
As our case studies illustrate, these \emph{conceptual causal models} clarify experimental logic, expose vulnerabilities, and guide further experiments toward more general, actionable insights.

Ultimately, adopting this pragmatic framework encourages a shift away from merely optimizing predictive performance on static benchmarks. By forcing us to make our structural assumptions explicit, causality provides the formal machinery to translate vague empirical choices into precise, testable hypotheses. This shift paves the way for more rigorous, transparent, and systematic progress not only in evaluating LLM reasoning, but across empirical machine learning.

\section*{Acknowledgment}

Felix Leeb is supported by the International Max Planck Research School for Intelligent Systems (IMPRS-IS). Zhijing Jin is supported by PhD fellowships from the Future of Life Institute and Open Philanthropy, as well as the travel support from ELISE (GA no 951847) for the ELLIS program. 

The material presented in this manuscript is partly based upon works supported by the German Federal Ministry of Education and Research (BMBF): Tübingen AI Center, FKZ: 01IS18039B; by the Machine Learning Cluster of Excellence, EXC number 2064/1 – Project number 390727645; 
the Swiss National Science Foundation (Project No. 197155); a Responsible AI grant by the Haslerstiftung; an ETH Grant (ETH-19 21-1); the Precision Health Initiative at the University of Michigan; and by the John Templeton Foundation (grant \#61156).







\bibliography{refs,example_paper}
\bibliographystyle{unsrtnat}


\clearpage
\appendix
\section{Examples of Monsters and Their Causal Interpretations}
\label{app:monsters}

The monsters in~\autoref{sec:monsters} are systematic, often hidden threats to the validity of benchmark conclusions. They describe recurring problems rather than mutually exclusive causal structures. Table~\ref{tab:monsters} connects nine such problems to concrete examples and candidate explanations. The terminology overlaps: one monster may admit several explanations, while the same CAT may help analyze several monsters. The interpretations below are modeling hypotheses; the cited symptom alone need not establish the proposed mechanism. Each interpretation motivates a diagnostic or intervention and makes clear what remains unresolved.

\begin{table}[ht]
\centering
\small
\setlength{\tabcolsep}{4pt}
\renewcommand{\arraystretch}{1.12}
\caption{Nine recurring monsters. Candidate causal patterns guide inquiry; they are neither exhaustive nor automatically identified by the examples. Further terminology, tests, and limitations appear below.}
\label{tab:monsters}
\begin{tabular}{@{}>{\raggedright\arraybackslash}p{0.25\linewidth}>{\raggedright\arraybackslash}p{0.32\linewidth}>{\raggedright\arraybackslash}p{0.38\linewidth}@{}}
\toprule
\textbf{Monster} & \textbf{Concrete example} & \textbf{Candidate causal pattern} \\
\midrule
Selection-induced social or cultural association & Western cultural defaults in Arabic-language evaluations & Selection can induce associations that subsequently influence model responses. \\
Perspective-dependent labels & Meaningful disagreement lost in a majority label & Perspective and context influence labels; aggregation determines the scoring target. \\
Label-conditioned annotation artifacts & NLI labels predictable from hypotheses alone & Requested label influences annotation choices and creates a cue used by the model. \\
Exposure-mediated leakage & Performance differs between familiar and fresh arithmetic benchmarks & Exposure provides an alternative pathway to success; exposure--accuracy associations may also be confounded. \\
Dataset-induced shortcut & Lexical overlap predicts entailment until controlled counterexamples & A dataset association becomes a causal influence on the trained model's predictions. \\
Pretraining-frequency or familiarity path & Arithmetic accuracy tracks term frequency & Familiarity may mediate a frequency effect; item properties may confound the observed association. \\
Unwanted sensitivity to task-irrelevant form & Reordering logically equivalent premises changes performance & Form influences the response despite an unchanged task and correct answer. \\
Benchmark-generator self-bias & Models score better on their own generated benchmarks & Generator-dependent domain, style, and labels influence measured performance. \\
Construct or population mismatch & One benchmark score stands in for a broader capability & The instrument measures a proxy, or selection changes the population to which conclusions apply. \\
\bottomrule
\end{tabular}
\end{table}

\subsection{Selection-induced social or cultural association}
\label{mon:selection}
\textit{Related terms: social bias, cultural bias, stereotyping, representational imbalance, selection bias.}
\citet{naousHavingBeerPrayer2024} find a preference for Western over Arab cultural entities and examine the composition of Arabic pretraining corpora. \citet{liPromptingFairnessIntegrating2025} explicitly use selection mechanisms to analyze social associations. If both social group and context affect inclusion in a corpus, conditioning on inclusion can induce an association between them. This motivates changes to selection/filtering criteria or information flow, evaluated across relevant populations. However, an observed cultural disparity does not by itself identify selection as its cause, and balancing frequencies does not settle which associations are appropriate for the task.

\subsection{Perspective-dependent labels}
\label{mon:perspective}
\textit{Related terms: label bias, subjectivity, disagreement, noise, perspectivism, cultural inequity.}
In their position paper, \citet{fleisigPerspectivistParadigmShift2024} examine how labeling practices treat disagreement and the assumptions behind attributing it to different perspectives. For a subjective judgment, majority aggregation can discard meaningful variation. A conceptual model separates the influences of perspective and supplied context on individual labels from the aggregation rule producing the scoring target. Preserving label distributions and reporting results by relevant perspectives can reveal dependence on those choices. The central question is whose judgments should be represented; causal analysis cannot resolve that normative choice, and demographic categories alone need not explain disagreement.

\subsection{Label-conditioned annotation artifacts}
\label{mon:annotation}
\textit{Related terms: annotation artifact, hypothesis-only bias, priming effect, label cue, label leakage.}
\citet{gururanganAnnotationArtifactsNatural2018} find class-predictive negation and vagueness in NLI datasets where workers generate a hypothesis for a specified entailment class. A hypothesis-only classifier can therefore succeed without comparing premise and hypothesis. The protocol creates a label-to-cue path; this need not be confounding. Changing the protocol, or editing cues while preserving the intended inference, can test this explanation. Adversarial filtering and counterfactual augmentation may reduce cue reliance, but filtering changes the evaluated population, and edits require checking that the correct label and relevant semantics remain intact.

\subsection{Exposure-mediated leakage}
\label{mon:leakage}
\textit{Related terms: benchmark leakage, contamination, train--test overlap, memorization, benchmark cheating, overfitting.}
\citet{zhouDontMakeYour2023} show how benchmark-related training can raise scores. \citet{zhangCarefulExaminationLarge2024} construct fresh arithmetic questions matched to GSM8K and find performance gaps for some models. Exposure can provide an alternative pathway to success through memorization, connecting this monster to mediation. Yet popularity or difficulty may also influence both exposure and accuracy. Private holdouts, temporal splits, and fresh matched questions reduce opportunities for leakage; controlled training interventions more directly test exposure effects. A performance gap alone does not prove contamination, since the compared questions may differ in other relevant ways.

\subsection{Dataset-induced shortcut}
\label{mon:shortcut}
\textit{Related terms: shortcut, heuristic, spurious correlation, competency problem, gaming.}
\citet{mccoyRightWrongReasons2019} identify lexical-overlap, subsequence, and constituent heuristics in NLI: models fail controlled examples where these heuristics conflict with the correct label. \citet{gardnerCompetencyProblemsFinding2021} discuss related single-feature dependencies as artifacts in datasets for competency problems. Training can turn a non-causal cue--label association into a causal influence of the cue on predictions. Contrast sets and validated edits can diagnose this reliance, motivating augmentation or regularization. However, the same feature may also carry legitimate information: mitigation should target the inappropriate dependence, rather than require invariance to every change in that feature.

\subsection{Pretraining-frequency or familiarity path}
\label{mon:frequency}
\textit{Related terms: frequency bias, familiarity effect, reciting, memorization, poor generalization.}
\citet{razeghiImpactPretrainingTerm2022} find that numerical-reasoning accuracy tracks the frequency of terms in pretraining data. One explanation is that frequency influences learned familiarity, which influences performance. Another is that item properties affect both frequency and difficulty. The observed correlation cannot distinguish these explanations. Matched sampling can address specified differences, while controlled training-frequency interventions more directly test the first pathway. Sampling evaluation items uniformly changes the evaluated distribution; it does not remove effects already learned during training. This monster therefore motivates both the confounding and mediation perspectives without establishing either from association alone.

\subsection{Unwanted sensitivity to task-irrelevant form}
\label{mon:form}
\textit{Related terms: brittleness, prompt sensitivity, premise-order bias, red herring, superficial variation, distractor sensitivity.}
\citet{chenPremiseOrderMatters2024} report performance drops after reordering logically equivalent premises; \citet{mirzadehGSMSymbolicUnderstandingLimitations2024} study unnecessary clauses in arithmetic problems. Such changes can affect responses even though the intended answer stays fixed. Randomized, semantics-preserving perturbations can estimate this behavioral effect with black-box access; no internal mechanism need be identified. The key assumption is that the intervention preserves the evaluated task. Changes that alter pragmatic interpretation, ambiguity, or relevant information require separate treatment, and robustness to one family of edits does not establish robustness to all forms of variation.

\subsection{Benchmark-generator self-bias}
\label{mon:selfbias}
\textit{Related terms: self-bias, self-enhancement, circular evaluation.}
\citet{yuanSilencerDiscoveryMitigation2025} find inflated performance on models' self-generated benchmarks and analyze contributions from question domain, language style, and incorrect labels. Generator identity thus affects the questions and reference answers through which performance is measured. Comparing generators, checking labels independently, and using heterogeneous generators can diagnose or reduce these pathways. This concerns benchmark construction, which is distinct from a judge favoring particular responses. Merely randomizing answer order or changing the judge does not address generator-dependent content. Residual differences in difficulty and coverage still need assessment against the intended evaluation target.

\subsection{Construct or population mismatch}
\label{mon:mismatch}
\textit{Related terms: construct validity, external validity, benchmark narrowness, task--ability mismatch, distribution shift, benchmark drift.}
\citet{rajiAIEverythingWhole2021,schlangenLanguageTasksLanguage2019,wallach2025position} question whether benchmark instruments adequately operationalize their intended capabilities. For example, success on a particular language task may be taken to establish a much broader ability. Separately, a benchmark population may differ from the deployment population. Measurement theory helps specify the construct and instrument; a causal model makes proposed mechanisms and population differences explicit. Evaluation across tasks and populations can test particular generalization claims, and transportability analysis can state the required assumptions~\citep{bareinboimTransportabilityCausalEffects2012}. Neither approach alone determines which capability should matter or guarantees that the chosen instrument captures it.





\end{document}